\documentclass[conference,a4paper]{IEEEtran}
\IEEEoverridecommandlockouts

\usepackage{fancyhdr}
\usepackage[dvips]{graphicx}
\usepackage{indent}
\usepackage{amsmath,amssymb}
\usepackage{paralist}
\usepackage{eclbkbox}
\usepackage{subfigure}

\makeatletter
\def\theequation{\arabic{equation}}
\def\tbcaption{\def\@captype{table}\caption}
\def\figcaption{\def\@captype{figure}\caption}
\makeatother

{\pointlessenum\begin{enumerate}}%
{\end{enumerate}}

\begin{document}
\title{Hierarchical Modular Reinforcement Learning Method and Knowledge Acquisition of State-Action Rule for Multi-target Problem
\thanks{\copyright 2013 IEEE. Personal use of this material is permitted. Permission from IEEE must be obtained for all other uses, in any current or future media, including reprinting/republishing this material for advertising or promotional purposes, creating new collective works, for resale or redistribution to servers or lists, or reuse of any copyrighted component of this work in other works.}
}

\author{\IEEEauthorblockN{Takumi Ichimura}
\IEEEauthorblockA{Faculty of Management and Information Systems,\\
Prefectural University of Hiroshima\\
1-1-71, Ujina-Higashi, Minami-ku,\\
Hiroshima, 734-8559, Japan\\
Email: ichimura@pu-hiroshima.ac.jp}
\and
\IEEEauthorblockN{Daisuke Igaue\authorrefmark{1}}
\IEEEauthorblockA{Graduate School of Comprehensive Scientific Research,\\
Prefectural University of Hiroshima\\
\authorrefmark{1} He graduated from Prefectural Univ. of Hiroshima\\ and is working at Iyo Bank, Ltd., Japan\\
Email: punch20@gmail.com}
}

\maketitle

\fancypagestyle{plain}{
\fancyhf{}	
\fancyfoot[L]{}
\fancyfoot[C]{}
\fancyfoot[R]{}
\renewcommand{\headrulewidth}{0pt}
\renewcommand{\footrulewidth}{0pt}
}

\pagestyle{fancy}{
\fancyhf{}
\fancyfoot[R]{}}
\renewcommand{\headrulewidth}{0pt}
\renewcommand{\footrulewidth}{0pt}

\begin{abstract}
Hierarchical Modular Reinforcement Learning (HMRL), consists of 2 layered learning where Profit Sharing works to plan a prey position in the higher layer and Q-learning method trains the state-actions to the target in the lower layer. In this paper, we expanded HMRL to multi-target problem to take the distance between targets to the consideration. The function, called `AT field', can estimate the interests for an agent according to the distance between 2 agents and the advantage/disadvantage of the other agent. Moreover, the knowledge related to state-action rules is extracted by C4.5. The action under the situation is decided by using the acquired knowledge. To verify the effectiveness of proposed method, some experimental results are reported.
\end{abstract}


\IEEEpeerreviewmaketitle

\section{Introduction}
\label{sec:Introduction}
Multi-Agent Systems (MAS) where there a number of autonomous agents interacting with each affecting the actions of the other agents is a complex system. Learning enables MAS to be more flexible and robust and makes agents better able to handle uncertain and changing circumstances. Thus how to coordinate the behaviors of different agents by learning method is required. Reinforcement learning is an area of machine learning in computer intelligent system \cite{Sutton98}, \cite{Grefenstette88}, \cite{Miyazaki99}. One of problems of reinforcement learning application of actual sized problem is ``curse of dimensional problem''. High dimension of input leads to huge number of rules in the reinforcement learning application.

In order to solve these problems several types of hierarchical reinforcement learning have been proposed to apply actual applications \cite{Wada09}, \cite{Watanabe10}. Hierarchical Modular Reinforcement Learning (HMRL), consists of 2 layered learning where Profit Sharing works to plan a prey position in the higher layer and Q-learning method trains the state-actions to the target in the lower layer. In this paper, we expanded HMRL to multi-target problem under the consideration of the distance between targets. The function, called `AT field', can estimate the interests for an agent according to the distance between 2 agents and the advantage/disadvantage of the other agent. Moreover, the knowledge related to state-action rules is extracted by C4.5. The action under the situation is decided by using the acquired knowledge. To verify the effectiveness of proposed method, some experimental results are reported.

The remainder of this paper is organized as follows. Section \ref{sec:ReinformcementLearning} describes about reinforcement learning method. Hierarchical modular reinforcement learning method is explained in Section \ref{sec:HierarchicalModularReinforcementLearning}. In the section, we explain the multi-agent pursuit problem. Moreover, we give consideration to deal with the value of target according to the distance between 2 prey agents. Section \ref{sec:KnowledgeAcquition} is the knowledge discover of learning agents in the format of If-Then rules. In Section \ref{sec:ConclusiveDiscussion}, we give some discussions to conclude this paper.

\section{Reinforcement Learning}
\label{sec:ReinformcementLearning}
The Profit Sharing and Q-Learning method are very popular in Reinforcement Learning. The section describes the algorithms of two kinds of Reinforcement Learning methods briefly. 

\subsection{Profit Sharing}
\label{sec:ProfitSharing}
Multi agent systems have been developed in the field of Artificial Intelligence. Each agent is designed to work some schemes based on many rules which indicate knowledge of the agent world or relationship among the agents. However, the knowledge or relationship is not always effective to survive in their environment, because the agent will discard a partial of knowledge if its environment changes dynamically. Reinforcement Learning \cite{Sutton98} is known to be worth to realize the cooperative behavior among agents even if little knowledge is provided with initial condition. The multi-agent system works to share a given reward among all agents.

 Especially, PS method \cite{Grefenstette88}, \cite{Miyazaki99} is an effective exploitation of reinforcement learning to adapt to a given environment. In PS, an agent learns a policy based on the reward that is received from the environment when it reaches a goal state. It is important to design a reinforcement function that distributes the received reward to each action rule in the policy. In PS, the rule $r_{i}$ is $(s, a)$ for possible action $a$ to a given sensory input $x$ to $s$. The rule ``If $x$ then $a$.'' is also written by $\overrightarrow{xa}$. PS does not estimate the value function and computes weight of rules $S_{r_{i}}$ for $(s, a)$. The episode is determined from the start state to the terminal state which the agent achieves the goal at time $i$ and then a reward ${\bf R}$ is provided. The PS gives the partial reward of ${\bf R}$ to the fired rule $(s_{i}, a_{i})$ in an episode($i<W$). $W$ is the maximum length of episode. The partial ${\bf R}$ is determined by the value function $f(i, {\bf R}, W)$. Each rule is reinforced by the sum of current weight and slanted reward. That is,
\begin{equation}
S_{r_{i}}=S_{r_{i}}+f_{i}, \: i=0, 1, \cdots, W-1,
\label{eq:profitsharing-1}
\end{equation}
where $S_{r_{i}}$ means the weight of the $i$th rule of an episode, $f_{i}$ is the reinforce function and means the reinforce value at the $-i$ step from obtaining ${\bf R}$.

\begin{figure}[btp]
\begin{center}
\includegraphics[scale=0.8]{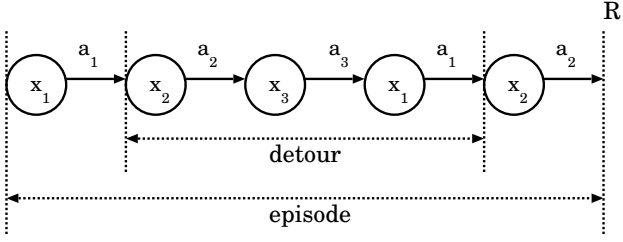}
\caption{The episode and the detour}
\label{fig:detour}
\vspace{-3mm}
\end{center}
\end{figure}

\begin{figure}[btp]
\begin{center}
\includegraphics[scale=1.0]{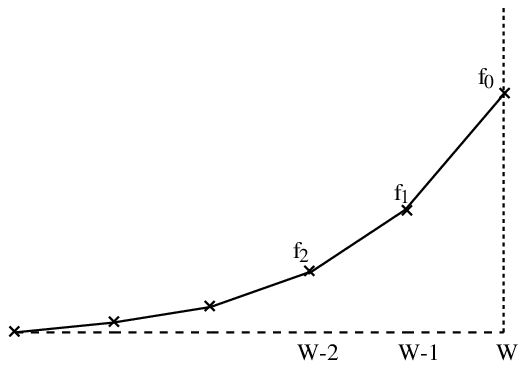}
\caption{Reinforcement Function}
\label{fig:reinforcementfunction}
\vspace{-3mm}
\end{center}
\end{figure}

The detour as shown in Fig.\ref{fig:detour} is the sequence of rules when the difference rules are selected for the same sensory input. There is a detour $(\overrightarrow{x_{2}a_{2}}, \overrightarrow{x_{3}a_{3}}, \overrightarrow{x_{1}a_{1}})$ in the sequence $(\overrightarrow{x_{1}a_{1}}, \overrightarrow{x_{2}a_{2}}, \overrightarrow{x_{3}a_{3}}, \overrightarrow{x_{1}a_{1}}, \overrightarrow{x_{2}a_{2}})$ in Fig.\ref{fig:detour}. The rules in the detour may occur some ineffective rules. The ineffective rule is always on the detour from the episode. The other rules are called the effective rule. If the competition between ineffective rules and effective rules exists, the ineffectiveness are not reinforced. If the reinforcement function satisfies the ineffective rule suppression theorem, the reinforcement function is able to distribute more reward to effective rules than ineffective ones. In order to suppress such ineffective rules, the forgettable PS method is proposed. 

\begin{equation}
L \sum_{j=1}^{w} f_{j} < f_{i-1}, \forall i= 1, 2, \cdots, W,
\label{eq:profitsharing-2}
\end{equation}
where $f_{i}$ is the reinforcement function and $L$ is the maximum number of effective rules. The reinforcement function decreases in a geometric series in the following.
\begin{eqnarray}
f_{i}=\frac{1}{M} f_{i-1}, i=1, 2, \cdots, W-1,
\label{eq:profitsharing-3}
\end{eqnarray}
where $M(\geq L+1)$ is a discount rate.
Eq.(\ref{eq:profitsharing-3}) reinforces the rule from $i=1$ to $i=W$ in an episode. Eq.(\ref{eq:profitsharing-3}) satisfied with the curve as shown in Fig. \ref{fig:reinforcementfunction}.

The algorithm of PS is as follows.

\begin{center}
\begin{indentation}{0.1cm}{0.1cm}
\begin{breakbox}
\smallskip
\begin{enumerate}[Step 1)]
\item Initialize $S_{r_{i}}$ arbitrarily.
\item Repeat (for each episode):
\begin{enumerate}
\item Initialize $r_{i}$ and $W$.
\item Repeat (for each step of episode):
\begin{enumerate}
\item $a \leftarrow$ action given by $\pi$ for $\mathcal{S}$ at state $x$ 
\item Take action $a$; observe reward, ${\bf R}$ and next state $\acute{x}$ 
\item $\forall i, i \leftarrow i+1$, set $r_{0}=\vec{x}_{a}$
\item If $R \neq 0$, set $f_{0}={\bf R}$ and calculate the following.
\begin{equation}
S_{r_{i}}=S_{r_{i}}+f_{i}, i=0, 1, \cdots, W-1,
\end{equation}
where $f_{i}=\frac{1}{M}f_{i-1}, i= 1, 2, \cdots, W-1$.
\item $x \leftarrow \acute{x}, W=W+1$
\end{enumerate}
\item until $x$ is terminal  
\end{enumerate}
\smallskip
\end{enumerate}
\end{breakbox}
\end{indentation}
\figcaption{The algorithm of PS}
\label{fig:algorithm-profitsharing}
\vspace{-5mm}
\end{center}

\subsection{$\mathcal{Q}$-Learning}
\label{sec:Q-Learning}
Temporal Difference (TD) method can directly learn from raw experience without a model of the environment's dynamics \cite{Sutton98}. TD method uses experience to solve the prediction problem. If a non-terminal state $s_{t}$ is visited at time $t$, TD method updates their estimate $V(s_{t})$ based on events after that visit. TD method waits only until the next time step. That is, TD method forms a target at time $t+1$ and makes an appropriate update using the observed reward $r_{t+1}$ and the estimate $V(s_{t+1})$. The simplest expression in TD method can be written as follows.

\begin{equation}
V(s_{t}) \leftarrow V(s_{t}) + \alpha \left[ r_{t+1} + \gamma V(s_{t+1}) - V(s_{t})\right]
\label{eq:TD-1}
\end{equation}

TD method can learn their estimates in part on the basis of other estimates. TD method is used for the evaluation or prediction by applying generalized policy iteration. We use the $\mathcal{Q}$-learning as an off-policy TD method in this paper, because the learned action-value function, $\mathcal{Q}$, directly approximates the optimal action-value function, $\mathcal{Q}^{*}$ with no dependence of policy. The simplest $\mathcal{Q}$-learning can be written as follows.

\begin{eqnarray}
\nonumber \mathcal{Q}(s_{t}, a_{t}) &\leftarrow& \mathcal{Q}(s_{t}, a_{t})\\
\nonumber && + \alpha \left[ r_{t+1} + \gamma \max_{a} \mathcal{Q}(s_{t+1},a) - \mathcal{Q}(s_{t},a_{t})\right]\\
\label{eq:Q-Learning-1}
\end{eqnarray}

\begin{center}
\begin{indentation}{0.1cm}{0.1cm}
\begin{breakbox}
\smallskip
\begin{enumerate}[Step 1)]
\item Initialize $\mathcal{Q}(s,a)$ arbitrarily.
\item Repeat (for each episode):
\begin{enumerate}
\item Initialize $s$.
\item Repeat (for each step of episode):
\begin{enumerate}
\item Choose $a$ from $s$ by using policy derived from $\mathcal{Q}$
\item Take action $a$; observe $r$ and next state $\acute{s}$
\item Eq.(\ref{eq:Q-Learning-1}) is executed.
\item $s \leftarrow \acute{s}$;
\end{enumerate}
\item until $x$ is terminal  
\end{enumerate}
\smallskip
\end{enumerate}
\end{breakbox}
\end{indentation}
\figcaption{The algorithm of $\mathcal{Q}$-Learning}
\label{fig:algorithm-QLearning}
\vspace{-3mm}
\end{center}

\section{Hierarchical Modular Reinforcement Learning Method}
\label{sec:HierarchicalModularReinforcementLearning}
This section defines Multi-Agent Pursuit Problem to explain the simulation environment where the Hierarchical Modular Reinforcement Learning (HMRL) \cite{Watanabe10} Method works. Moreover, we develop the HMRL method to work in Multi-Agent Pursuit Problem where two or more kinds of prey agents works in the same environment.

\subsection{Multi-Agent Pursuit Problem}
\label{sec:Multi-AgentPursuitProblem}
\begin{figure}[btp]
\begin{center}
\includegraphics[scale=0.35]{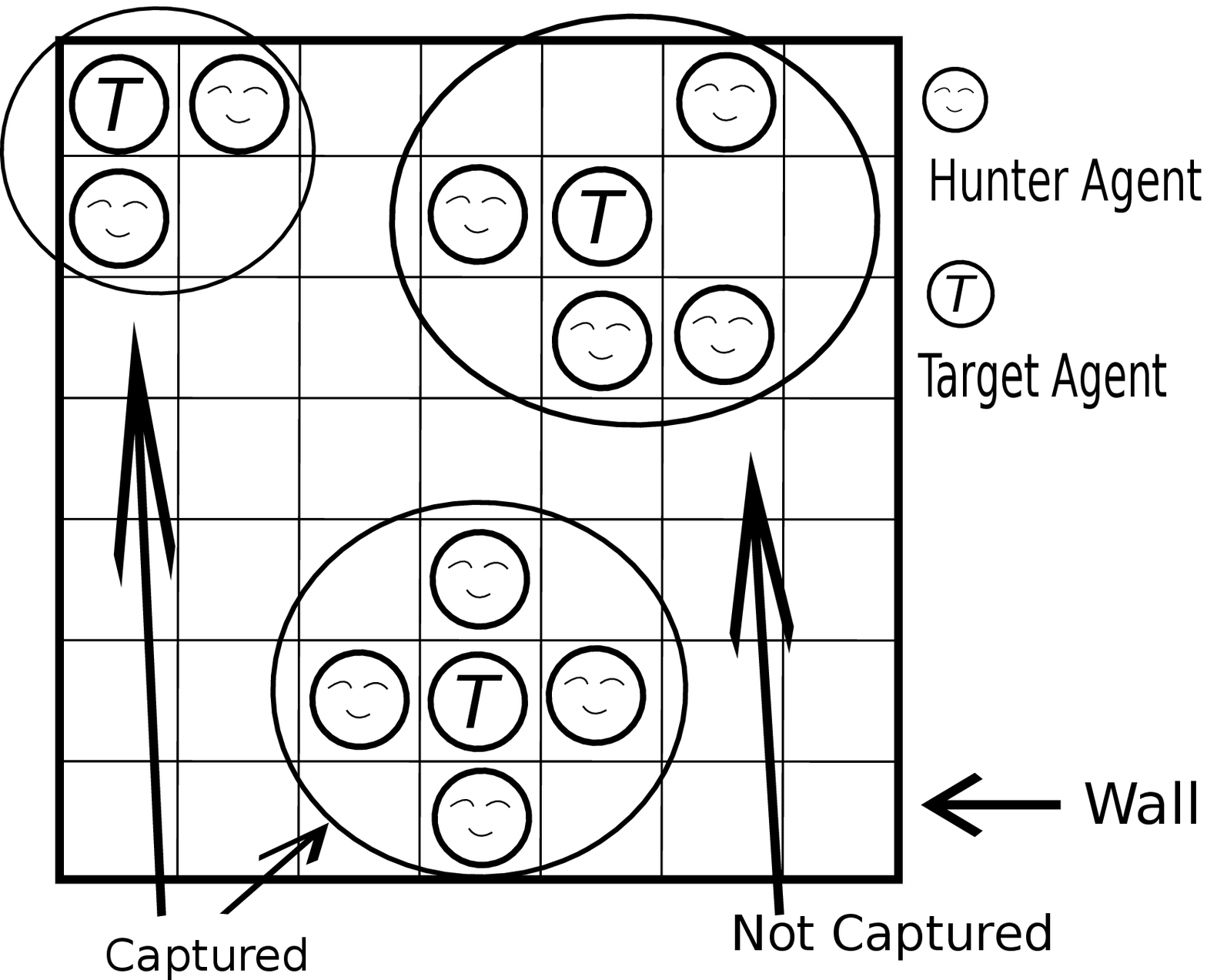}
\caption{Multi-Agent Pursuit Problem}
\vspace{-3mm}
\label{fig:Multi-AgentPursuitProblem}
\vspace{-3mm}
\end{center}
\end{figure}

The pursuit problem is well-known to be an appropriate example of cooperative Mulit-agent system (MAS) \cite{Pu-Cheng06}. In this study, the pursuit problem is considered in a $7 \times 7$ grid world, where two prey agents($T$) and four hunter agents are placed at random positions in the environment as shown in Fig.\ref{fig:Multi-AgentPursuitProblem}. Hunters are learning agents and try to capture the randomly moving prey. In this paper, the prey agent does not learn the state-action rule through the experience and the two or more prey agents does not work to cooperate with each other. At each time, agents synchronously select and perform on out of five actions without communicating with each other: Staying at the current position or moving north, south, west, or east. Preys and hunters cannot share a cell. Also, an agent is not allowed to move off the environment. The prey is captured, when all of its neighbor cells are occupied by hunters as shown in Fig.\ref{fig:Multi-AgentPursuitProblem}.

\subsection{Hierarchical Modular Reinforcement Learning Method}
\label{sec:HRL}
For the pursuit problem, huge memory consumption is required to express the internal knowledge of the agents. Moreover, because the surrounding environment is complex, the agents cannot express the collaboration. \cite{Wada09}, \cite{Watanabe10} proposed the hierarchical modular reinforcement learning to solve the above problems. It is difficult to decide how many kinds of sub-task should be decomposed into. 

In \cite{Watanabe10}, Prof. Watanabe conceived of the idea that decomposes the surrounding task(capturing) into ``decision of move position target'' for surrounding according to current monitored state and ``selection of appropriate action'' to move to the target position of each agent. The task is decomposed into ``surrounding'' task synchronized with the other hunter agents and ``exploring the environment'' task. Moreover, the upper task corresponds only to collaborative surrounding strategy.

\begin{figure}[btp]
\begin{center}
\includegraphics[scale=0.5]{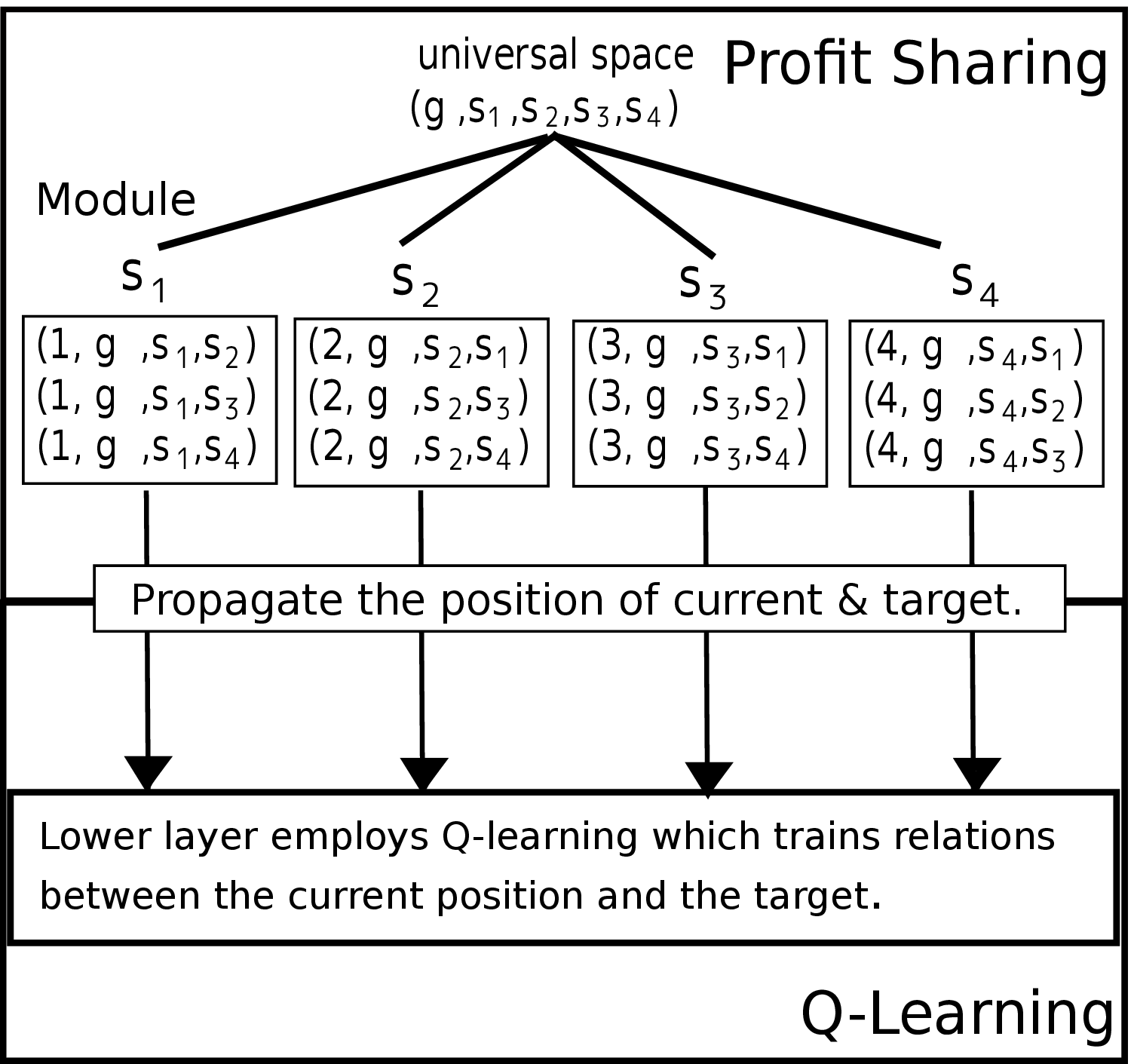}
\caption{Hierarchical Reinforcement Learning}
\label{fig:HRL_Model}
\vspace{-3mm}
\end{center}
\end{figure}
\begin{figure}[btp]
\begin{center}
\includegraphics[scale=0.5]{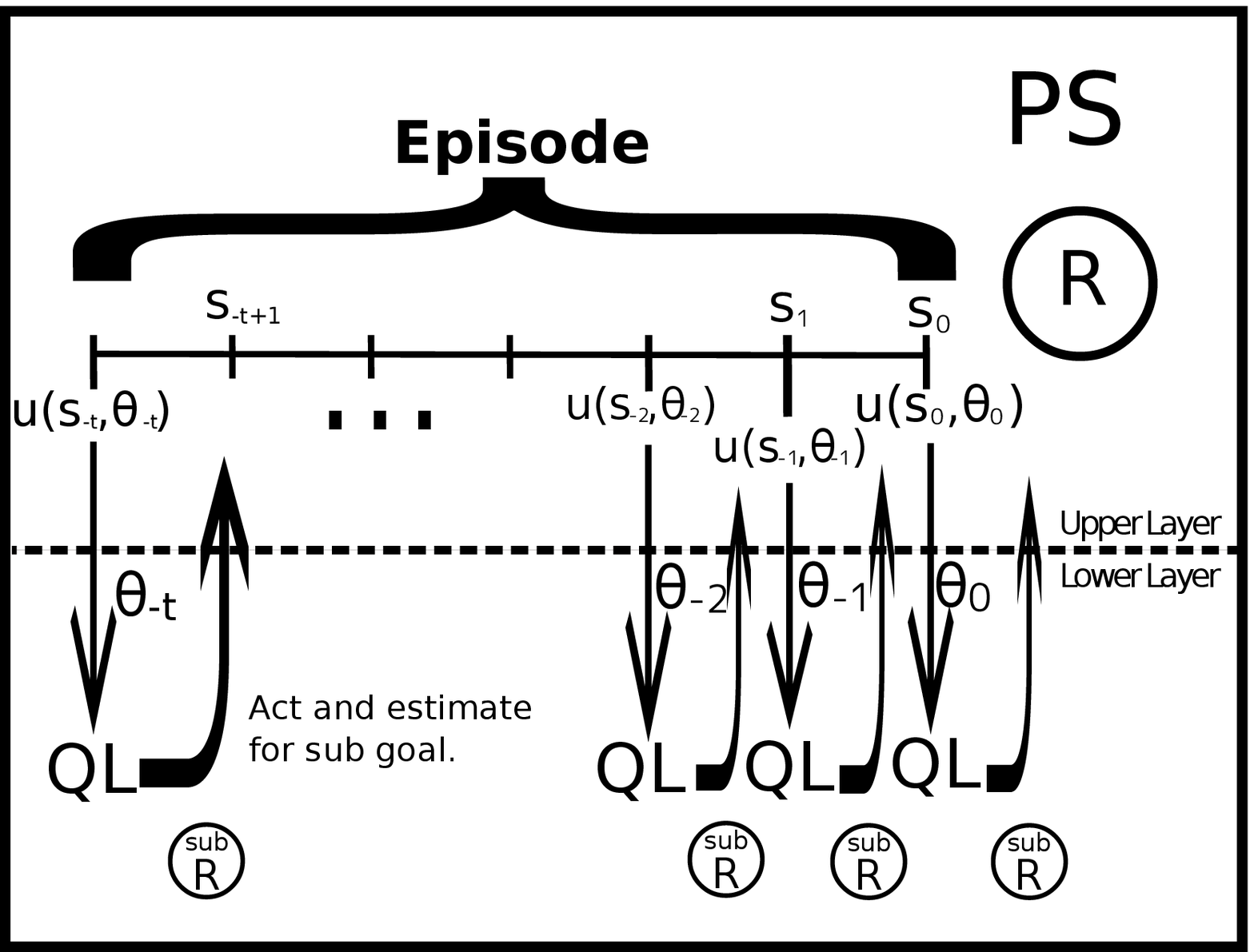}
\vspace{-3mm}
\caption{A Distribution of Reward in Hierarchical Reinforcement Learning}
\label{fig:HRL_flow}
\vspace{-3mm}
\end{center}
\end{figure}

In the upper layer, the target position of the agent is decided based on observed state such as the current position of the prey agent and the other hunter agents. The rules in the upper layer express goodness of the target position corresponding to the current state excluding actual actions. In order to construct the rules based on the current state combination, huge corresponding memory is needed. To avoid such requirement, the authors applied modular structure for the rule expression \cite{Watanabe10} in the upper layer as shown in Fig.\ref{fig:HRL_Model}. In Fig.\ref{fig:HRL_Model}, the state space is divided to 4 sub-spaces where the following equation is satisfied.
\begin{equation}
(g,s_{1},s_{2},s_{3},s_{4})=\cup_{e}(e,g,s_{e},s_{\epsilon}), (e,\epsilon\in E,e\neq\epsilon)
\label{eq:space_division}
\end{equation}

The weights of rules in the upper layer are updated by Profit Sharing as follows.
\vspace{-3mm}
\begin{eqnarray}
\nonumber & u(e,g(i),h_{e}(i),h_{\epsilon}(i))=u(e,g(i),h_{e}(i),h_{\epsilon}(i))\\
\nonumber & +k(e,g(i),h_{e}(i),h_{\epsilon}(i)),\\
\nonumber & k(e,g(i-1),h_{e}(i-1),h_{\epsilon}(i-1))=\\
\nonumber & \rho k(e,g(i),h_{e}(i),h_{\epsilon}(i))\\
& (i=0,-1,\cdots,-m,\epsilon\neq e),
\label{eq:upper_update}
\end{eqnarray}
where $u(\cdot)$ is the estimate function for target position and $k(\cdot)$ is an reinforcement function as shown Fig.\ref{fig:reinforcementfunction}. $e$ is the hunter agent and $\epsilon$ is the other hunter agent. $g(i)$ is the position and  $h_{e}(i)$ is the position at time $i$, respectively. Time $0$ is when the hunter agent receives the reward. $\rho$ is the parameter.

The target position is divided as a sub goal for surrounding tasks instead of final goal corresponding to the current state of the prey agent according to the weight of rules. The target position of the agent is determined by the following equation.
\vspace{-3mm}
\begin{eqnarray}
\nonumber \theta_{e}=arg\max_{v}\sum_{\epsilon}\frac{u(e,g,v,h_{\epsilon})}{\mu^{|h_{e}-v|}}, \,
(\epsilon\neq e,\mu\geq 1),
\label{eq:upper_theta}
\end{eqnarray}
where $v$ is the candidate of target position. According to the selected state, the information for target position is sent to the lower layer.

In the lower layer, the selection of action to walk to the target position decided at the upper layer is implemented by reinforcement learning process as Q-learning:
\begin{eqnarray}
\nonumber &Q(s_{e}(t),a_{e}(t),\theta_{e})=Q(s_{e}(t),a_{e}(t),\theta_{e})\\
\nonumber &+k(r_{t}+\gamma \max_{\eta}Q(s_{e}(t+1),\eta,\theta_{e})-Q(s_{e}(t),a_{e}(t),\theta_{e})), 
\label{eq:lower_q}
\end{eqnarray}
where $Q$ is $Q$-value, $s_{e}(t)$ and $a_{e}(t)$ are the state vector and the action of the agent $e$ at $t$th step, respectively. $\theta_{e}$ is the position of agent $e$. $r_{t}$ is the reward. $\eta$ is the maximum value of action. $k$ is the step size parameter.

\subsection{2 prey agent based Hierarchical Reinforcement Learning}
\label{sec:HRL_2TA}

\begin{figure}[btp]
\begin{center}
\includegraphics[scale=0.3]{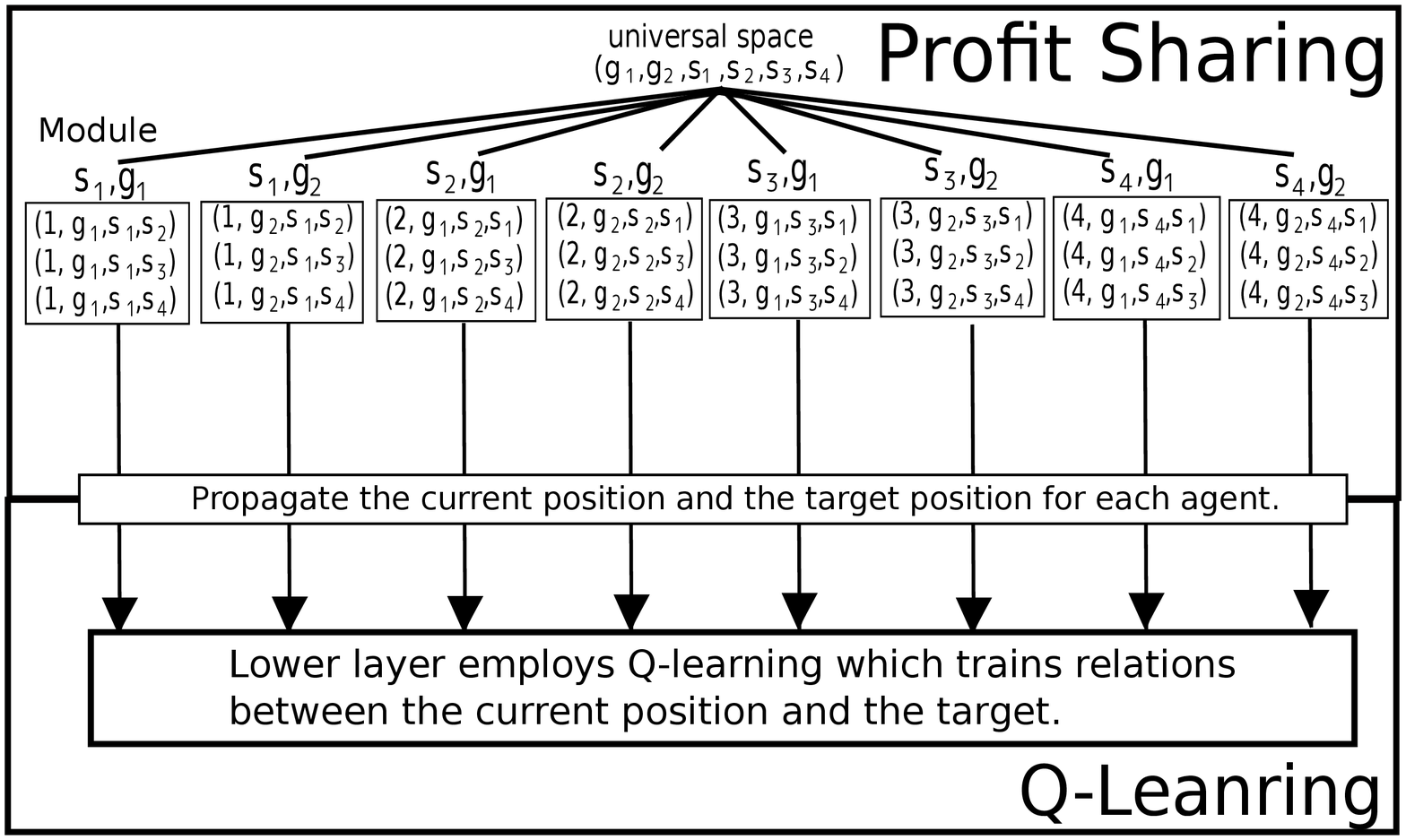}
\caption{2 Prey Agent Model}
\label{fig:2PA_Model}
\vspace{-3mm}
\end{center}
\end{figure}

Multi-Agent Pursuit Problem with 2 prey agents in the environment is discussed in this paper. For the problem, we consider the division of space as shown in Fig.\ref{fig:2PA_Model}. If there are 2 prey agents, the environment has 2 goals. Therefore, the relation among sub spaces in Fig.\ref{fig:2PA_Model} is defined as follows:
\vspace{-3mm}
\begin{eqnarray}
\nonumber (g_{1},g_{2},s_{1},s_{2},s_{3},s_{4})=\cup_{e}\cup_{l}(e,g_{l},s_{e},s_{\epsilon})\\
(e,\epsilon\in E,l\in L,e\neq\epsilon),
\label{eq:space_division_2agent}
\end{eqnarray}
where $g_{l}$ is the goal position for each prey agent.

Each modular has 2 target position, but only one target position should be sent to the lower layer. Therefore, the judgment rule for the decision of appropriate position is defined as follows:
\vspace{-3mm}
\begin{eqnarray}
\nonumber \theta_{e}=\left\{
{\large
\begin{array}{l}
arg \max_{v}\sum_{\epsilon}\frac{u(e,g_{0},v,h_{\epsilon})}{\mu^{|h_{e}-v|}}\\
\nonumber \indent if\; |h_{e}-g_{0}|<|h_{e}-g_{1}|\\
arg \max_{v}\sum_{\epsilon}\frac{u(e,g_{1},v,h_{\epsilon})}{\mu^{|h_{e}-v|}}\\
\nonumber \indent if\; |h_{e}-g_{1}|<|h_{e}-g_{0}|\\
\end{array}
}
\right.\\
(\epsilon\neq e,\mu\geq 1)
\label{eq:targetposition_2agent}
\end{eqnarray}

If the target is quite different altering behaviors among the prey agents, e.g. a target has positive reinforcement and the other agent has punishment, it is difficult to consider the value of target simultaneously. In this paper, Eq.(\ref{eq:ATField}) is defined by the idea that when there are 2 kinds of target due to their value, positive and negative, the value of $u()$ is changed according to the degree to be affected by each other.
\begin{equation}
ATF=\left\{
\begin{array}{lll}
\Phi&=&0.0\:(if\;gd\leq n_{1})\\
\Phi&=&1.0\:(if\;n_{1}<gd\leq n_{2})\\
\Phi&=&0.9\:(if\;n_{2}<gd)
\end{array}
\right. ,\\
\label{eq:ATField}
\end{equation}
where $gd$ is the distance between two agents. $n_{1}$ is the parameter to judge for whether the distance of the agent and the other agent is within close distance and $n_{2}$ is the parameter to judge whether the distance is long distance. In this paper, we set that $n_{1}=2$ and $n_{2}=5$. The estimate value is updated by using Eq.(\ref{eq:ATField_update}).
\vspace{-3mm}
\begin{eqnarray}
\label{eq:ATField_update}
&u(e,g_{l}(i),h_{e}(i),h_{\epsilon}(i))=u(e,g_{l}(i),h_{e}(i),h_{\epsilon}(i))\nonumber\\
&+k(e,g_{l}(i),h_{e}(i),h_{\epsilon}(i))\nonumber\\
&k(e,g_{l}(i-1),h_{e}(i-1),h_{\epsilon}(i-1))\nonumber\\
&= \rho\cdot ATF(gd)\cdot k(e,g_{l}(i),h_{e}(i),h_{\epsilon}(i))\\
&(e,\epsilon\in E,l\in L,i=0,-1,\cdots,-m), \nonumber
\end{eqnarray}
where $ATF(gd)$ is the function of AT-Field given by Eq.(\ref{eq:ATField}). The output of function can be reduced by the discount factor, $\rho$, according to the degree that the corresponding agent is affected by the other agent. $e$ and $\epsilon$ are the index of hunter and prey agent, respectively. $E$ and $L$ mean the set of all agents and the prey agent, respectively. $h$ and $g$ are the position of the hunter agent and the prey agent, respectively. $ATF$ function will not affect the division of state space in the profit sharing. 

\subsection{Simulation Results}
This section describes the simulation results under the 2 prey agent and 4 hunter agents. The position of all agents are randomly assigned in the $7 \time 7$ grid. A trial is starting from the initial situation until the hunter agents capture 2 prey agents as shown in Fig.\ref{fig:Multi-AgentPursuitProblem}. After one trial the environment and $Q$-value are initialized, and a set of simulation is till 20,000 trials. The reward is 100 if the prey agent is positive target and it is 0 otherwise. In lower layer, when the agent reaches to the target position sent from the upper layer, the agent can receive the reward 100. The behavior of prey agent is randomly and the hunter agent moves due to the acquired state-action rules. Of course, each agent does not know the behaviors of the other agents.

In order to evaluate the effectiveness of the proposed model, we define the 3 ratio of capturing targets: ``(Within Safety)'', ``(Within Dangerous)'', and ``(Positive Ratio)''.

\begin{enumerate}
\item (Within Safety): When the hunter agent captured a prey agent with positive reward, if the distance between them is larger than $n_{1}$, the captured prey agents belongs to the set $far$. 
\begin{eqnarray}
P({\rm safety\_distance})=\frac{\# ({\rm safety\_target}\cap {\rm far})}{\# ({\rm safety\_target})}
\end{eqnarray}
\item (Within Dangerous): When the hunter agent captures a prey agent with positive reward, if the distance between them is smaller than $n_{1}$, the captured prey agents belongs to the set $near$. 
\begin{eqnarray}
P({\rm dangerous\_distance})=\frac{\# ({\rm safety\_target} \cap {\rm near})}{\# ({\rm safety\_target})}
\end{eqnarray}
\item (Positive Ratio): It means the ratio of (Within Safety) over the simulations. 
\begin{eqnarray}
P({\rm safety\_target_{positive}})=\frac{\# ({\rm safety\_target} \cap {\rm far})}{{\rm iteration}}
\end{eqnarray}
\end{enumerate}

Table \ref{tab:step_withATF} and Table \ref{tab:step_woATF} show that the number of steps and the actions without ATField model and with one, respectively, until the prey target is captured. The simulation result related to the number of steps and actions are almost same results, although the computation time with ATField model gets longer than that without ATField model.

Table \ref{tab:capture_withATF} and Table \ref{tab:capture_woATF} show that the capture ratio of (Safety Target), (Within Safety), (Within Dangerous), (Positive Ratio), and distance, without ATField model and with one, respectively. The distance means the distance between targets. From these tables, the performance in model with ATField is better than that without ATField model.

\begin{table}[btp]
\begin{center}
\caption{Step Number without ATF}
\begin{tabular}{|c|c|c|c|c|} \hline
Iterations & \multicolumn{2}{|c|}{Episode}& \multicolumn{2}{|c|}{Action}\\ \cline{2-5}
&Ave.&Var.&Ave.&Var.\\ \hline
1-200& 684.7 & 2374.0& 958.6& 1596.7\\ \hline
201-2,000& 355.2 & 127.9& 358.5& 134.5\\ \hline
2,001-17,000 & 101.3 & 5.1& 104.9& 5.3\\ \hline
17,001-20,000 & 62.3 & 3.3 & 66.6& 3.1\\ \hline
\end{tabular}
\label{tab:step_withATF}
\vspace{-3mm}
\end{center}
\end{table}

\begin{table}[tbp]
\begin{center}
\caption{Step Number with ATF}
\begin{tabular}{|c|c|c|c|c|} \hline
Iterations & \multicolumn{2}{|c|}{Episode}& \multicolumn{2}{|c|}{Action}\\ \cline{2-5}
&Ave.&Var.&Ave.&Var.\\ \hline
1-200 & 703.5 & 2881.1& 999.1& 2372.9 \\ \hline
201-2,000& 403.7 & 193.4 & 406.7& 203.7\\ \hline
2,001-17,000 & 119.7 & 4.2 & 122.7& 4.6\\ \hline
17,001-20,000 & 74.5 & 3.7& 77.7& 3.6\\ \hline
\end{tabular}
\label{tab:step_woATF}
\vspace{-3mm}
\end{center}
\end{table}

\begin{table*}[btp]
\begin{center}
\caption{Capture Ratio of Prey Agents without ATF}
\begin{tabular}{|c|c|c|c|c|c|c|c|c|c|c|} \hline
&\multicolumn{2}{|c|}{Safety Target} & \multicolumn{2}{|c|}{Within Safety} & \multicolumn{2}{|c|}{Within Dangerous} & \multicolumn{2}{|c|}{Positive ratio}&\multicolumn{2}{|c|}{Distance} \\ \cline{2-11}
&Ave.&Var.&Ave.&Var.&Ave.&Var.&Ave.&Var.&Ave.&Var.\\ \hline
1-200 & 53.1\%& 14.4  &60.2\%&　24.0 &39.8\% &　24.0 & 32.0\%& 14.0 &3.22& 0.030\\ \hline
201-2000 & 74.8\% & 1.4 & 77.2\%& 2.7 &22.8\%& 2.7 & 57.8\% & 4.0&3.95& 0.004\\ \hline
2001-17000 & 86.4\% & 0.2 & 84.3\%& 0.1 &15.7\% & 0.1 & 72.9\% & 0.3 & 4.27& 0.001\\ \hline
17001-20000 & 84.7\%  & 0.7 & 83.3\% & 0.4&16.7\%& 0.4& 70.6\%& 0.9 & 4.11& 0.001\\ \hline
\end{tabular}
\label{tab:capture_withATF}
\vspace{-3mm}
\end{center}
\end{table*}

\begin{table*}[btp]
\begin{center}
\caption{Capture Ratio of Prey Agents with ATF}
\begin{tabular}{|c|c|c|c|c|c|c|c|c|c|c|} \hline
&\multicolumn{2}{|c|}{Safety Target} & \multicolumn{2}{|c|}{Within Safety} & \multicolumn{2}{|c|}{Within Dangerous} & \multicolumn{2}{|c|}{Positive ratio}&\multicolumn{2}{|c|}{Distance} \\ \cline{2-11}
&Ave.&Var.&Ave.&Var.&Ave.&Var.&Ave.&Var.&Ave.&Var.\\ \hline
1-200& 53.7\% & 24.5 &49.2\% &　13.9 &50.8\% &　13.9 &26.5\%& 16.6 &2.93 & 0.03\\ \hline
201-2000& 73.4\% & 1.2 & 77.2\% & 2.7 &22.8\% & 2.7 &56.7\% & 2.7 &3.94 & 0.004\\ \hline
2001-17000 & 89.7\% & 0.1 & 86.9\% & 0.04 &13.1\% & 0.04 &77.9\% & 0.1 & 4.51 & 0.0005\\ \hline
17001-20000 & 90.6\% & 0.4 & 86.4\%& 0.5 &13.6\% & 0.5 &78.3\% & 0.3 & 4.41 & 0.001\\ \hline
\end{tabular}
\label{tab:capture_woATF}
\vspace{-3mm}
\end{center}
\end{table*}

\section{Knowledge Acquisition}
\label{sec:KnowledgeAcquition}
The state-action rules in the lower layer are extracted by C4.5. Fig.\ref{fig:decision_tree_result} shows the part of extracted results. The rules are extracted while training the module. Fig.\ref{fig:decision_tree_result} is the result of 19,900-20,000 trials. `theta\_x' and `theta\_y' mean the difference in the x axis and y axis in the move, respectively. The output in the teaching signal is the target position sent from the upper layer. The simulation is 72,327 instances in the teach data set. 
\begin{center}
\begin{indentation}{0.1cm}{0.1cm}
\begin{breakbox}
\tiny
\begin{quote}
{\begingroup54pt \lineskiplimit=-\maxdimen
\begin{verbatim}
theta_Y > -1
|   theta_X <= -1
|   |   theta_Y <= 0: left (12519.0/1907.0)
|   |   theta_Y > 0
|   |   |   theta_Y <= 1: left (2172.0/1096.0)
|   |   |   theta_Y > 1
|   |   |   |   theta_X <= -2
|   |   |   |   |   theta_Y <= 2: down (270.0/128.0)
|   |   |   |   |   theta_Y > 2
|   |   |   |   |   |   theta_X <= -3
|   |   |   |   |   |   |   theta_X <= -5
|   |   |   |   |   |   |   |   theta_Y <= 3: left (4.0/1.0)
|   |   |   |   |   |   |   |   theta_Y > 3: down (2.0/1.0)
|   |   |   |   |   |   |   theta_X > -5
|   |   |   |   |   |   |   |   theta_Y <= 3: down (20.0/8.0)
|   |   |   |   |   |   |   |   theta_Y > 3: stay (5.0/2.0)
|   |   |   |   |   |   theta_X > -3: left (56.0/29.0)
|   |   |   |   theta_X > -2: down (648.0/328.0)
|   theta_X > -1
|   |   theta_Y <= 0
|   |   |   theta_X <= 0: stay (8056.0)
|   |   |   theta_X > 0
|   |   |   |   theta_X <= 2: right (12260.0/1733.0)
|   |   |   |   theta_X > 2
|   |   |   |   |   theta_X <= 3: right (320.0/179.0)
|   |   |   |   |   theta_X > 3
|   |   |   |   |   |   theta_X <= 4: stay (442.0/104.0)
|   |   |   |   |   |   theta_X > 4: right (47.0/33.0)
|   |   theta_Y > 0
|   |   |   theta_X <= 0
|   |   |   |   theta_Y <= 1: down (11541.0/1451.0)
|   |   |   |   theta_Y > 1
|   |   |   |   |   theta_Y <= 2: down (959.0/352.0)
|   |   |   |   |   theta_Y > 2
|   |   |   |   |   |   theta_Y <= 3: down (328.0/199.0)
|   |   |   |   |   |   theta_Y > 3
|   |   |   |   |   |   |   theta_Y <= 5: stay (173.0/83.0)
|   |   |   |   |   |   |   theta_Y > 5: left (11.0/4.0)
\end{verbatim}
\endgroup}
\end{quote}
\end{breakbox}
\end{indentation}
\figcaption{Calculation Results of C4.5 (partial)}
\label{fig:decision_tree_result}
\end{center}

For easy comprehension, Fig.\ref{fig:State-ACtion_rules} shows the extracted knowledge as shown in Fig.\ref{fig:decision_tree_result} in the If-Then rule format. In the simulation, we can get 47 state-action rules. Fig.\ref{fig:State-ACtion_rules} shows 10 sample rules only.

\begin{center}
\begin{indentation}{0.1cm}{0.1cm}
\begin{breakbox}
\tiny
\begin{quote}
{\begingroup
\baselineskip=5pt \lineskiplimit=-\maxdimen
\begin{verbatim}
No.1
If theta_X <= 4 theta_X > 2 theta_Y <= -6 Then up 
with CF=1.0
No.2
If theta_X <= 0 theta_X > -1 theta_Y <= 0 theta_Y > -1 Then stay 
with CF=1.0
No.3
If theta_X <= 0 theta_X > -1 theta_Y <= 1 theta_Y > 0 Then down 
with CF=0.8742743263148774
No.4
If theta_X <= 2 theta_X > 0 theta_Y <= 0 theta_Y > -1 Then right 
with CF=0.8586460032626427
No.5
If theta_X <= 0 theta_X > -1 theta_Y <= -1 Then up 
with CF=0.8478816513050886
No.6
If theta_X <= -1 theta_Y <= 0 theta_Y > -1 Then left 
with CF=0.8476715392603243
No.7
If theta_X <= 4 theta_X > 3 theta_Y <= 0 theta_Y > -1 Then stay 
with CF=0.7647058823529411
No.8
If theta_X <= -5 theta_Y <= 3 theta_Y > 2 Then left CF=0.75
No.9
If theta_X <= 1 theta_X > 0 theta_Y <= 5 theta_Y > 4 Then stay 
with CF=0.7272727272727273
No.10
If theta_X <= -6 theta_Y <= -1 theta_Y > -2 Then left 
with CF=0.7142857142857143
\end{verbatim}
\endgroup}
\end{quote}
\end{breakbox}
\end{indentation}
\figcaption{IF-Then rules (partial)}
\label{fig:State-ACtion_rules}
\vspace{-3mm}
\end{center}

By using the acquired rules, the simulation results are the number of steps and the actions, and the ratio of captured prey agents as shown in Table \ref{tab:step_knowledge} and \ref{tab:capture_knowledge}. The performance with rules are better than that of without rules.

\begin{table}[tbp]
\begin{center}
\caption{Step Number}
\begin{tabular}{|c|c|c|} \hline
&Episode&Action\\ \hline
1-200& 543.1 & 543.4 \\ \hline
201-2,000 & 194.0 & 195.6\\ \hline
2,001-17,000 & 63.0 & 69.3\\ \hline
17,001-20,000 & 46.3 & 54.6\\ \hline
\end{tabular}
\label{tab:step_knowledge}
\vspace{-3mm}
\end{center}
\end{table}

\begin{table}[tbp]
\begin{center}
\caption{Capture Ratio of Prey Agent}
\begin{tabular}{|c|p{1cm}|p{1cm}|p{1cm}|p{1cm}|p{1cm}|} \hline
& Safety Target & Within Safety & Within Dangerous & Positive ratio & Target Distance  \\ \hline
1-200 & 56.5\%  &61.1\% &38.9\% &34.5\% &3.27\\ \hline
201-2,000 & 85.7\% & 84.8\% &15.2\% &72.6\% &4.44\\ \hline
2,001-17,000 & 92.2\% & 85.5\% &14.6\% &78.9\%& 4.36\\ \hline
17,001-20,000 & 91.4\%  & 84.3\% &15.7\% &77.1\%& 4.23\\ \hline
\end{tabular}
\label{tab:capture_knowledge}
\vspace{-3mm}
\end{center}
\end{table}

\section{Conclusive Discussion}
\label{sec:ConclusiveDiscussion}
Hierarchical Modular Reinforcement Learning (HMRL)\cite{Watanabe10}, consists of 2 layered learning where Profit Sharing works to plan a prey position in the higher layer and Q-learning method trains the state-actions to the target in the lower layer. If the multi-agent pursuit problem has 2 or more prey agents, in many cases, the reward for them is set toward same purpose, that is, the rewards are same value. In this paper, we expanded HMRL to multi-target problem under the consideration of the distance between targets. The function, called `AT field', can estimate the interests for an agent according to the distance between 2 agents and the advantage/disadvantage of the other agent. Moreover, the knowledge related to state-action rules is extracted by C4.5. In simulation results, AT field function is effective to measure the difference between the rewards of prey agents. We will verify the method in real world problem in future.

\end{document}